\documentclass[conference]{IEEEtran}
\IEEEoverridecommandlockouts    

\usepackage{caption}
\usepackage[dvips]{graphicx}
\usepackage{amssymb}
\usepackage{amsmath}
\usepackage{cite}
\usepackage{color}
\usepackage{xcolor}
\usepackage{subfig}
\usepackage{subfloat} % @#@
\usepackage{enumitem}
\usepackage{algorithm}
\usepackage{comment}
\usepackage{tabularx}
\usepackage{enumitem}
\usepackage{tabularx}
\usepackage{graphicx}
\usepackage{glossaries}
\usepackage{listings}
\usepackage{multirow}
\usepackage{booktabs}
\usepackage{mathtools}
\usepackage{ragged2e}
\usepackage{tabularx}
\usepackage{tikz}
\usepackage{schemabloc}
\usepackage{siunitx}
\usepackage{stackengine}
\usepackage{verbatim}
\usepackage{adjustbox}
\usepackage{algpseudocode}
\usepackage{placeins}

\usepackage{siunitx}
\usepackage{booktabs}
\usepackage{array,xcolor,colortbl,booktabs}
\usepackage{xcolor}

\def\BibTeX{{\rm B\kern-.05em{\sc i\kern-.025em b}\kern-.08em
    T\kern-.1667em\lower.7ex\hbox{E}\kern-.125emX}}

\title{\LARGE \bf \vspace{17pt}
Enhancing Sim2Real Transfer for Torque-Controlled Robots through Real2Sim Dynamics Estimation and Reinforcement Learning
}

\author{Davide Bargellini$^{1}$ Alex Pasquali$^{1}$ Andrea Govoni$^{1}$ Riccardo Zanella$^{2}$ and Gianluca Palli$^{1}$% <-this % stops a space
\thanks{$^{1}$D. Bargellini, A. Pasquali, A. Govoni, and G. Palli are with the Department of Electrical, Electronic, and Information Engineering (DEI), University of Bologna, 40136 Bologna, Italy.}
\thanks{$^{2}$R. Zanella is with the Robotics and Mechatronics (RaM) group, University of Twente, Enschede, The Netherlands.}%
\thanks{This work is supported by the Horizon Europe project IntelliMan-AI-Powered Manipulation System for Advanced Robotic Service, Manufacturing and Prosthetics [grant number 101070136].}%
}

\begin{document}

\maketitle

\begin{abstract}
Transferring reinforcement learning policies from simulation to Real-World robots remains a major challenge, particularly when dealing with low-level torque control, where even small modelling inaccuracies can lead to unstable or unsafe behaviours. In this work, we propose a Real2Sim2Real pipeline that improves Sim2Real transfer for torque-controlled robotic arms by combining trajectory matching, parameter optimization via genetic algorithms, and domain randomization. Using the 7-DOF Franka Emika Panda robot, we first identify friction, inertia, and gravity compensation parameters by minimizing the error between real and simulated joint trajectories. These calibrated dynamics are then used to train a TQC-based reinforcement learning agent in simulation. The trained policy is evaluated in both Gazebo and MuJoCo environments, and finally deployed on the real robot. Our results demonstrate a significant improvement in tracking accuracy and policy robustness after parameter tuning, with smooth policy transfer from simulation to the Real-World across multiple target-reaching tasks. This work highlights the effectiveness of accurate physical modelling in enabling stable and generalizable torque-based reinforcement learning policies.
\end{abstract}

\section{Introduction}\label{s:Introduction}

Torque control has emerged as a promising approach to improve interaction with the environment, since it enables more precise responses to physical forces and constraints \cite{ROVEDA2021103711}, \cite{RLImpedanceControl}. However, training directly in the real world with \textit{joint torques} is time-consuming, resource-intensive, and safety-critical. Simulations mitigate these limits by enabling faster and safer policy development before hardware deployment.

However, a key challenge arises when these control policies, developed in the simulated environment, are transferred to Real-World robots \cite{simtorealDeepRLSurvey}, \cite{diarel}. This issue is commonly referred to as the Sim2Real transfer problem\cite{UserGuide}. Despite the advantages of simulation, there are often discrepancies between simulated and Real-World environments, such as differences in physical parameters or contact forces, and between different simulators themselves \cite{JaumeTiagoRL}. Each simulator may implement physics and models differently, leading to variations in the results. These discrepancies create a "reality gap" \cite{RealityGap}, where control policies that perform well in simulation may fail to generalize in Real-World scenarios, leading to a decline in performance.

To mitigate this gap, recent studies have focused on improving the physical accuracy of simulations through better calibration of parameters such as dynamics, friction, and sensor feedback \cite{DynamicsRandomization}, \cite{AutoTuneSim2Real}, \cite{offline_domain_randomization}.

Our study focuses on optimizing the Sim2Real transfer, particularly emphasizing the development of a Reinforcement Learning (RL) agent using different policies such as DDPG\cite{DDPG}, SAC\cite{SAC}, TD3\cite{TD3}, and TQC\cite{TQC}, for the 7-axis Franka Emika Panda robot that utilizes joint torques to reach a position. This preliminary work aims to fill the gap in torque control before the robot interacts with the environment.
The paper is organized as follows: Sec \ref{s:Related_Works} summarizes the related works, while Sec. \ref{s:Methodologies} provides a detailed overview of the methodology and the mathematical tools used. Finally, Sec. \ref{s:E_R} presents the experiments and results of the Sim2Real transfer, and Sec. \ref{s:Conclusions} discusses the conclusions drawn from the developed work and outlines future directions.

\begin{figure*}[t]
    \vspace*{1mm}
    \centering
    \includegraphics[width = 1.0\linewidth]{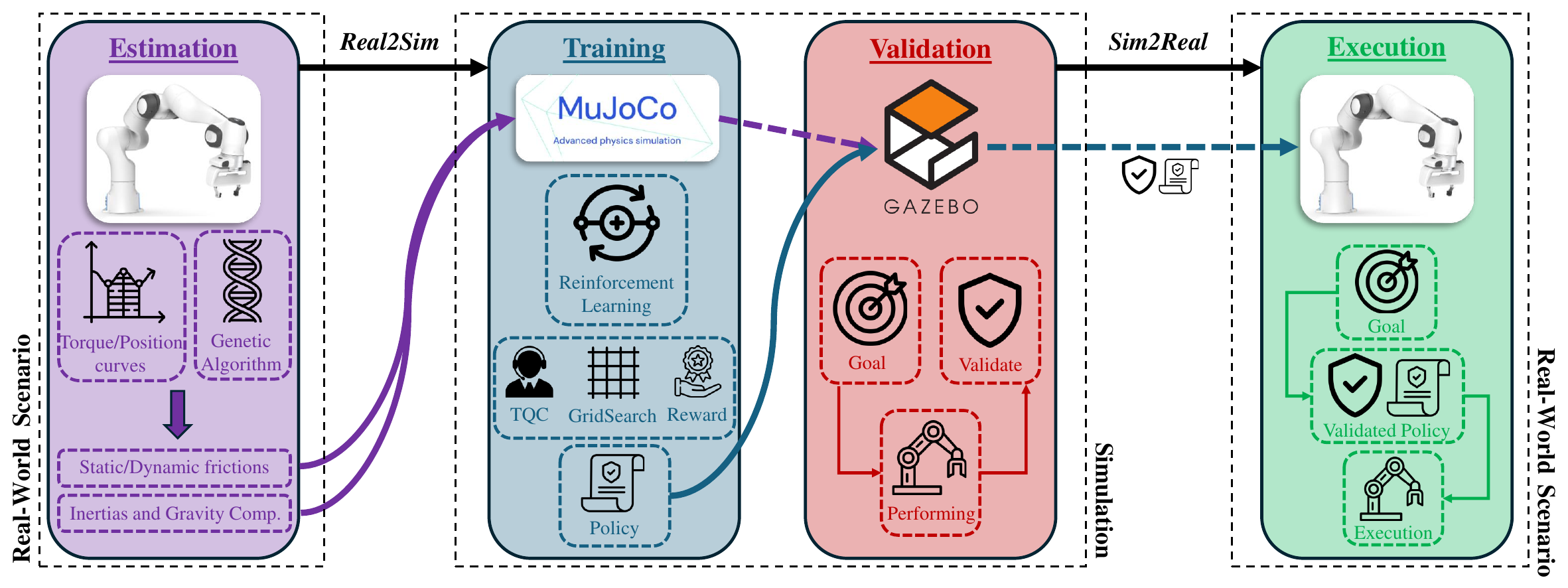}
    \caption{\small Diagram illustrating the workflow for validating and executing an RL policy in both simulation and Real-World scenarios. The process includes parameter estimation (e.g., torque/position curves, friction, inertias, and gravity compensation), training via RL (TQC with GridSearch optimization), policy validation, and execution for Real2Sim and Sim2Real applications.}
    \label{fig:scheme}
\end{figure*}
\section{Related Works}\label{s:Related_Works}
Training robotic agents in joint space is a promising approach for addressing complex robot-environment interactions, as demonstrated in \cite{DynamicsRandomization}, \cite{DRLJointTorque}. This method provides a deeper understanding of how the robot interacts with its surroundings and allows the exploitation of full-model dynamics. As a result, it can lead to improved performance while requiring less domain-specific knowledge, especially when dealing with complex floating-base robots with discrete contact dynamics.\\
However, the effectiveness of an agent trained in simulation heavily depends on how accurately the physical parameters of the robot and its environment are modelled. The discrepancy between simulated and Real-World parameters, known as the reality gap, can hinder the transfer of learned policies from simulation to the Real-World (Sim2Real transfer). Therefore, accurate parameter estimation is crucial to improve the alignment between simulation and reality, and consequently, the success of the agent's training.

One of the most common approaches to tackle the reality gap is domain randomization (DR) \cite{DynamicsRandomization}, \cite{offline_domain_randomization}, which introduces controlled variability in simulated physical parameters, such as mass, friction, inertia, and material resistance, during agent training. This method aims to make the agent robust to uncertainties in the Real-World, allowing the robot to learn from more scenarios and achieve better results in reality.\\
Another approach is trajectory matching \cite{Sim2RealAdpatSimWithRealData}, \cite{Sim2RealwSim}, which attempts to bridge the reality gap by using Real-World data to create simulations with dynamics that closely resemble those observed in practice.

RL algorithms enable robots to adapt and learn in dynamic, uncertain environments. They can optimize control strategies for motor torques based on feedback, improving control accuracy and efficiency, and ultimately enhancing the robot’s performance across a wide range of tasks and environments \cite{DRLJointTorque}.\\
One of the challenges in using RL with torque control is the complexity of modelling motor dynamics, especially when considering interactions between motors and their impact on the robot's movement. Designing control algorithms that can effectively manage motor torques requires a deep understanding of the system's dynamics.

The challenge we address is how to accurately encode realistic dynamics by reducing the gap between simulation and reality, aiming to achieve robot behaviour in simulation that closely reflects Real-World dynamics through careful fine-tuning of parameters, particularly friction and inertia, using trajectory matching and genetic algorithm \cite{GA1}, \cite{GA2}. Friction parameters are calculated based on \cite{IdFrankaParam}, while inertia parameters are optimized using trajectory matching, through a genetic optimization algorithm discussed in detail in Sec. \ref{ss:Parameters}. Once these initial parameters are estimated, the RL agent is trained over a domain randomized around that value. This approach aims to ensure more robust performance in Real-World tasks.

\section{Methodologies}\label{s:Methodologies}
This section presents the methodology used to achieve torque-level control on robotic manipulators. We first reduce the Sim2Real gap by calibrating key dynamics parameters via Real2Sim trajectory matching, using a genetic algorithm to minimize the discrepancy between measured and simulated joint responses. The resulting calibrated simulator is then used to train a control policy entirely in simulation, improving the fidelity of policy transfer to the real robot.

\subsection{Reinforcement Learning Overview}\label{ss:RL}
In robotics, RL is often used to learn control policies directly from interaction, which is convenient when torque-level behaviours must emerge from complex robot–environment dynamics. An agent interacts with an environment, chooses actions from its current state, receives reward feedback, and updates its behaviour to maximize long-term cumulative reward through exploration and exploitation.
In this work, RL's agent is a 7-DoF Franka Emika Panda arm. To obtain a policy that outputs joint torque commands. The policy is trained to reach Cartesian target positions in the workspace while maintaining smooth and stable motion in transfer from simulation to the real robot.

The RL agent is trained in simulation, using techniques such as DR to ensure robust performance when transferring the learned control policies to Real-World scenarios. The agent's goal is mathematically represented as maximizing the expected cumulative reward, defined as:

\begin{equation}
    R_t = \sum_{k=0}^{\infty} \gamma^k r_{t+k},
\end{equation}

where \( R_t \) is the expected cumulative reward at time \( t \), \( r_{t+k} \) is the reward received at time \( t+k \), and \( \gamma \) is the discount factor, which determines the importance of future rewards.

The reward function \( r \) for the RL agent, in our environment, is expressed as:
\begin{equation}
\begin{split}
r = & -k_1 d - k_2 \sum_{i=1}^{7} \tanh \left( |\dot{\theta}_{i}| \right) - \\
    & \frac{k_3 \sum_{i=1}^{7} \tanh \left( |\ddot{\theta}_{i}| \right)}{|k_4 - \min(1, d)|} - 
    \frac{k_5 \sum_{i=1}^{7} \tanh \left( |\ddot{\theta}_{i}| \right)}{k_6 + d},
\end{split}
\end{equation}

Here, \(k_{1},...,k_{6}\) are positive scaling constants. In our settings: \(k_1=2\)  weights distance-to-goal, \(k_2=0.03\) weights velocity penalty, \(k_3=0.05\) weights acceleration penalty, and \(k_4=1.15, k_5=0.03, k_6=0.15\) are small positive constants used in the acceleration denominators. The normalized distance $d$ is evaluated keeping the maximum distance as the norm between the end effector's initial position and the target.  \(\dot{\theta}_i\) and \(\ddot{\theta}_i\) denote joint velocity and acceleration; \(\tanh\) bounds their contributions to encourage smooth motions.

The reward function is designed to balance accuracy in reaching the goal position with smooth and energy-efficient joint movements. The first term encourages proximity to the goal, while the velocity and acceleration penalties (via $\dot{\theta}_i$ and $\ddot{\theta}_i$) discourage abrupt or unstable motion. The use of $\tanh$ ensures numerical stability and limits the influence of extreme values.

Through iterative training, the agent learns to associate specific joint torques with successful task completion, gradually improving its policy \( \pi \) to minimize errors and enhance task efficiency. The policy update can be expressed as:

\begin{equation}
    \Theta_{t+1}=\Theta_t-\eta(t) \nabla L,
\end{equation}

where \( \Theta_{t} \) represents the parameters of the policy at time \( t \), \( \eta(t) \) is the learning rate, \( L(\Theta) \) is the loss function and \(\nabla L \) denotes the gradient of the loss\cite{wu2023selecting}.

This approach fundamentally relies on tuning the reward function and simulation parameters to replicate Real-World dynamics closely.

\subsection{The Parameter Estimation Problem}\label{ss:Parameters}

The effectiveness of RL in robotic control is heavily influenced by the accuracy of the model parameters that define the robot's dynamics. In this section, we focus on the parameter estimation problem.\\
To describe the robot’s dynamics, we consider the standard equation of motion for an $n$-DOF manipulator:

\begin{equation}\label{eq:robot_model}
\tau = M(q)\ddot{q} + C(q, \dot{q})\dot{q} + G(q) %+ F(\dot{q})
\end{equation}

where $\tau$ is the vector of joint torques, $M(q)$ is the inertia matrix, $C(q, \dot{q})$ represents Coriolis friction and centrifugal effects, and $G(q)$ is the gravity vector. In simulation, inertial and geometric parameters are taken from the robot description file. Instead of using standard friction and damping coefficients from the robot description file, they are modelled separately, following the identified parameters in \cite{Barge, IdFrankaParam}, and include directional asymmetry, i.e., different friction levels for positive and negative joint velocities.
For each joint $j$, the (velocity-dependent) dynamic friction torque is computed as:

\begin{equation}\label{eq:dynamic_friction}
\tau_{f, j}^{dyn} = \frac{\varphi_{1, j}}{1 + e^{-\varphi_{2, j}(\dot{q}_j + \varphi_{3, j})}} - \frac{\varphi_{1, j}}{1 + e^{-\varphi_{2, j} \varphi_{3, j}}}, \quad j \in [1, \ldots, 7]
\end{equation}

Where \(\tau_{f, j}\) is the friction compensation torque, \(\varphi_{1, j}\), \(\varphi_{2, j}\), and \(\varphi_{3, j}\) are parameters specific to each joint to be estimated. 
Static friction is not identified by the velocity-dependent model in eq.\eqref{eq:dynamic_friction} and is instead introduced manually as a joint-wise term. In practice, we define a dead-zone around zero velocity, i.e., $|\dot{q}_j| < \dot{q}_{\min}$, within which a constant breakaway torque $\tau^{\mathrm{st}}_{j}$ (tuned empirically for each joint) is applied with sign given by the desired motion direction:
\begin{equation}
\tau^{\mathrm{st}}_{f,j} =
\begin{cases}
\tau^{\mathrm{st}}_{j}\,\mathrm{sgn}(\dot{q}^{\mathrm{des}}_j), & \text{if } |\dot{q}_j| < \dot{q}_{\min},\\[2mm]
0, & \text{otherwise,}
\end{cases}
\qquad j\in\{1,\ldots,7\}.
\end{equation}
% The overall friction compensation is then obtained as
% \begin{equation}
% \tau_{f,j} = \tau^{\mathrm{dyn}}_{f,j}(\dot{q}_j) + \tau^{\mathrm{st}}_{f,j}.
% \end{equation}

The estimation problem for these parameters is based on minimizing the error between the simulated and Real-World trajectories, which is expressed as:

\begin{equation}
\label{eq:gen_minimization}
    E(\mathbf{p}) = \sum_{t=0}^{T} \left\| \mathbf{y}_{\text{sim}}(t, \mathbf{p}) - \mathbf{y}_{\text{real}}(t) \right\|,
\end{equation}

where \( E(\mathbf{p}) \) is the error function, \( \mathbf{y}_{\text{sim}}(t, \mathbf{p}) \) is the simulated trajectory based on the parameter set \( \mathbf{p} \), \( \mathbf{y}_{\text{real}}(t) \) is the corresponding Real-World trajectory, and \( T \) is the total time of observation. Minimizing this error ensures that the simulated dynamics closely replicate the Real-World behaviour, serving as the basis for successful policy transfer. Each optimization iteration improves the fidelity of simulated joint trajectories.

To bridge the simulation-to-reality gap, we adopted a genetic algorithm approach~\cite{GA1} that minimizes trajectory tracking error by iteratively adjusting the parameter set \( \mathbf{p} \) for each joint, which includes the previously mentioned parameters in eq.\eqref{eq:robot_model}. The algorithm calibrates the simulated joint dynamics to better match Real-World behaviour, improving Sim2Real transfer of the learned control policies. A genetic algorithm iteratively tunes the parameters by minimizing an error metric, yielding a robust estimate that enhances the RL agent’s performance in both simulation and on the real robot (see Alg.~\ref{alg:genetic})

% \begin{figure}[!b]
% \centering
% \subfigure[Operator's hand near the robot hand with repulsion gradients]{
%     \includegraphics[scale=0.35]{images/3.A/Joint1_torque1.png}
%     \label{fig:img1}} 
% \hspace{0.4cm}
% \subfigure[Point camera noise on the robot surface filtered out; actual collision detected is the operator hand]{
%     \includegraphics[scale=0.35]{images/3.A/Joint1_torque-2.png}
%     \label{fig:img2}}
% \vspace*{-3mm}
% \caption{Collision detection in presence of operator hand and surface noise.}
% \label{fig:collision_operator}
% \vspace*{-3mm}
% \end{figure}

\begin{algorithm}
\caption{\small Genetic Algorithm for Parameter Estimation}
\label{alg:genetic}
\begin{algorithmic}[1]

\State Sample a real trajectory $\mathbf{y}_{\text{real}}(t)$
\State \textbf{Initialize} population $\mathcal{P} = \{\mathbf{p}_1, \mathbf{p}_2, \dots, \mathbf{p}_N\}$
\While{not converged \textbf{and} generation $< G_{\text{max}}$}
    \For{each $\mathbf{p}_i \in \mathcal{P}$}
        \State Simulate trajectory $\mathbf{y}_{\text{sim}}(t, \mathbf{p}_i)$
        \State Compute error using Eq.~\ref{eq:gen_minimization}
    \EndFor
    \State Select top-performing individuals based on $E(\mathbf{p}_i)$
    \State Apply mutation to generate new individuals
    \State Update population $\mathcal{P}$
\EndWhile
\State \Return best parameter set $\mathbf{p}^*$
\end{algorithmic}
\end{algorithm}

\section{Experiments and Results} \label{s:E_R}

% \begin{figure}[!t]
% \centering
% \subfigure[Far noise; no active link, no repulsion triggered]{
%     \includegraphics[scale=0.35]{images/3.A/Joint1_torque1_new.png}
%     \label{fig:img3}} 
% \hspace{0.4cm}
% \subfigure[Another scenario description here]{
%     \includegraphics[scale=0.35]{images/3.A/Joint1_torque-2_new.png}
%     \label{fig:img4}}
% \vspace*{-3mm}
% \caption{Collision filtering examples under various spatial noise scenarios.}
% \label{fig:collision_noise}
% \vspace*{-3mm}
% \end{figure}

\begin{figure}[t]
    \centering
    \includegraphics[width = \columnwidth]{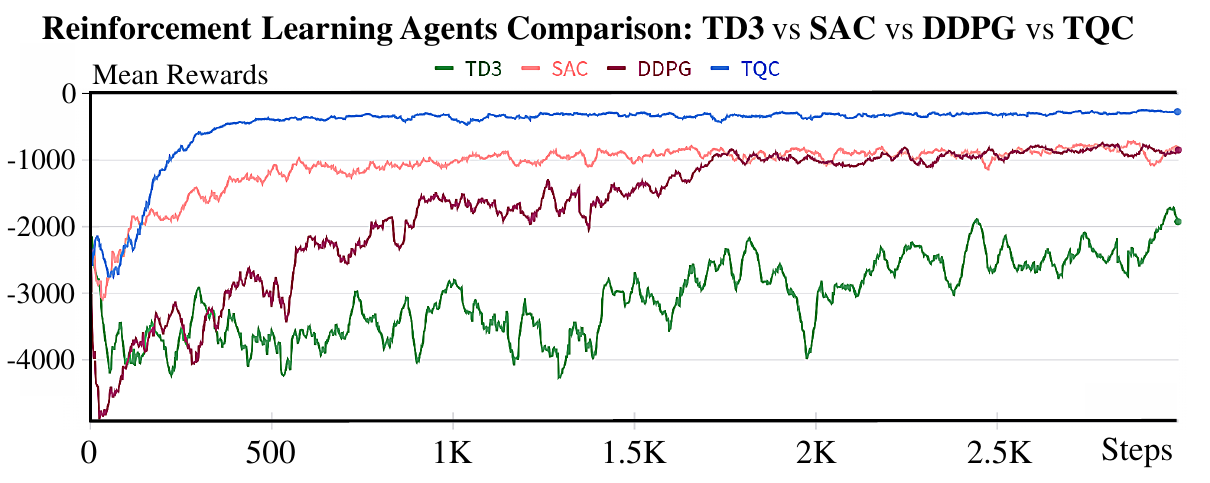}
    \caption{\small Episode mean reward comparison across DDPG, SAC, TD3, and TQC algorithms. TQC demonstrates consistently higher rewards, underscoring its superior performance and robustness in our environment.}
    \label{fig:TQCcomparison}
    \vspace*{-4mm}
\end{figure}

We evaluated DDPG, SAC, TD3, and TQC, tuning each method via grid search. Overall, TQC provided the most robust and stable performance across metrics (Fig.~\ref{fig:TQCcomparison}) and showed lower sensitivity to hyperparameters, which is desirable for Sim2Real deployment. For this reason, we report the Sim2Real transfer results using TQC.

\subsection{Parameters Estimation}\label{ss:ParameterEstimation}

\begin{figure}[t]
    \centering
    \subfloat[]{\includegraphics[width=0.49\columnwidth]{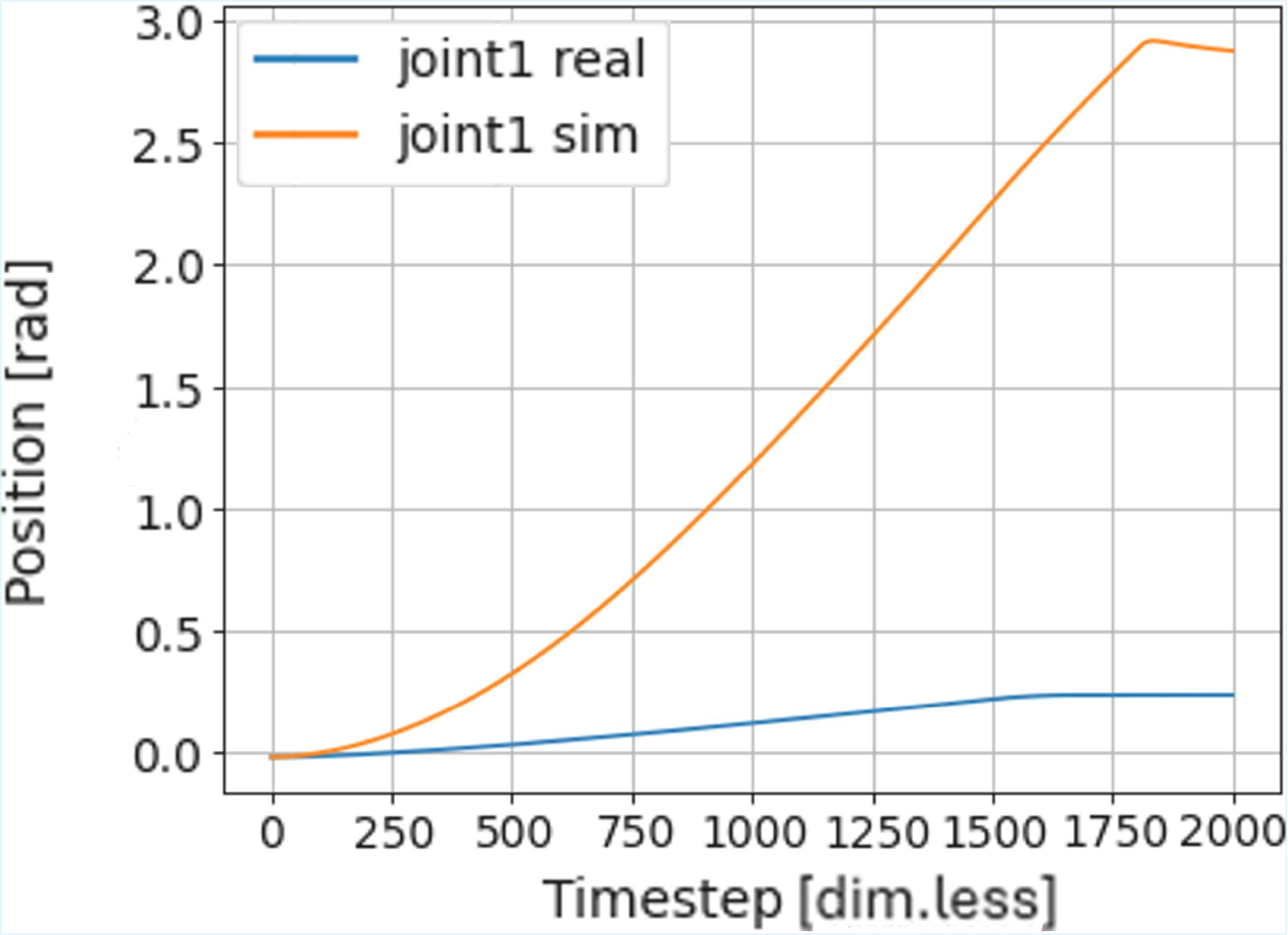} \label{fig:J1NoTune}}
    \subfloat[]{\includegraphics[width=0.49\columnwidth]{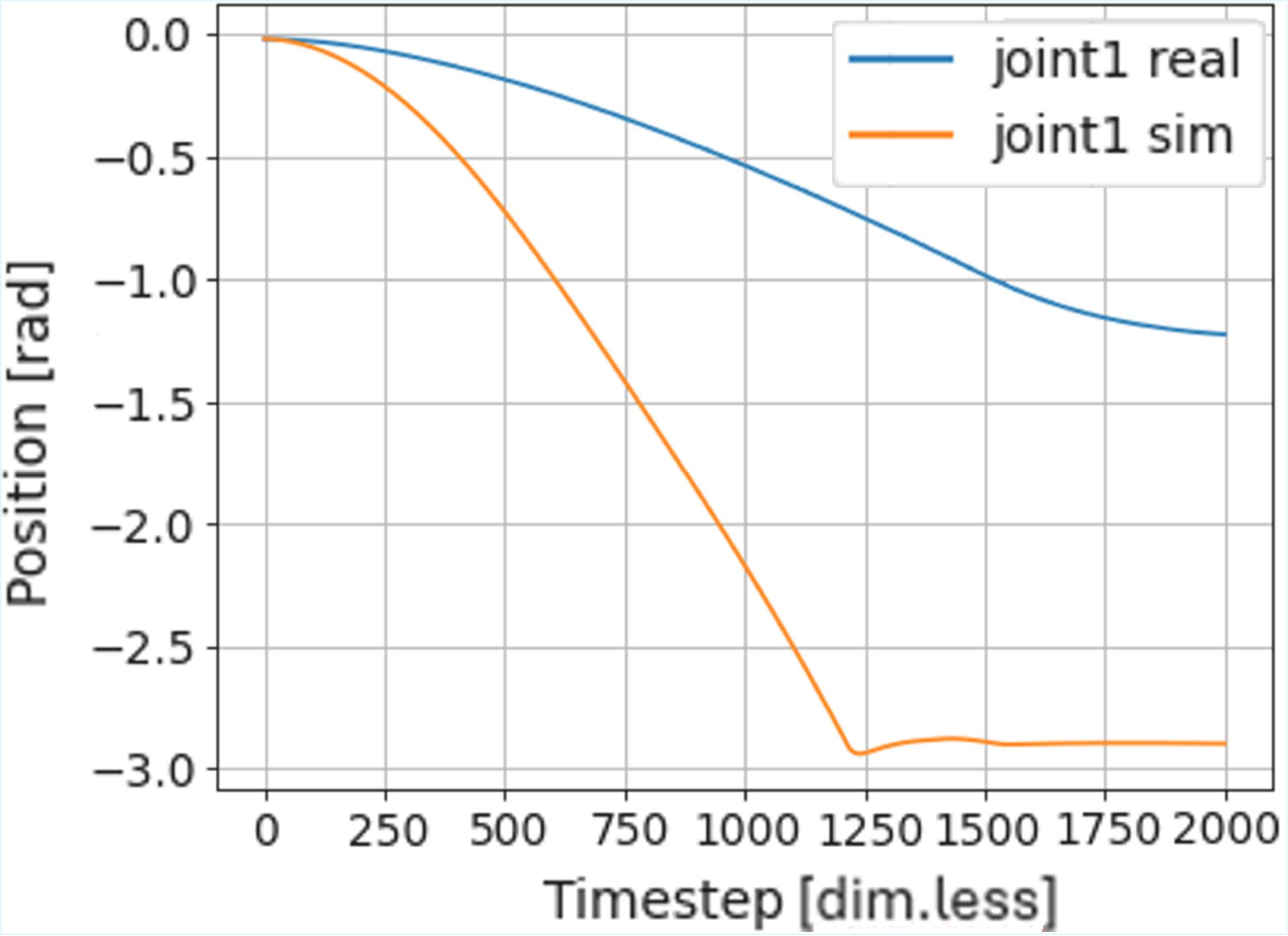} \label{fig:J2NoTune}}
    \vspace{-3mm}
    
    \subfloat[]{\includegraphics[width=0.49\columnwidth]{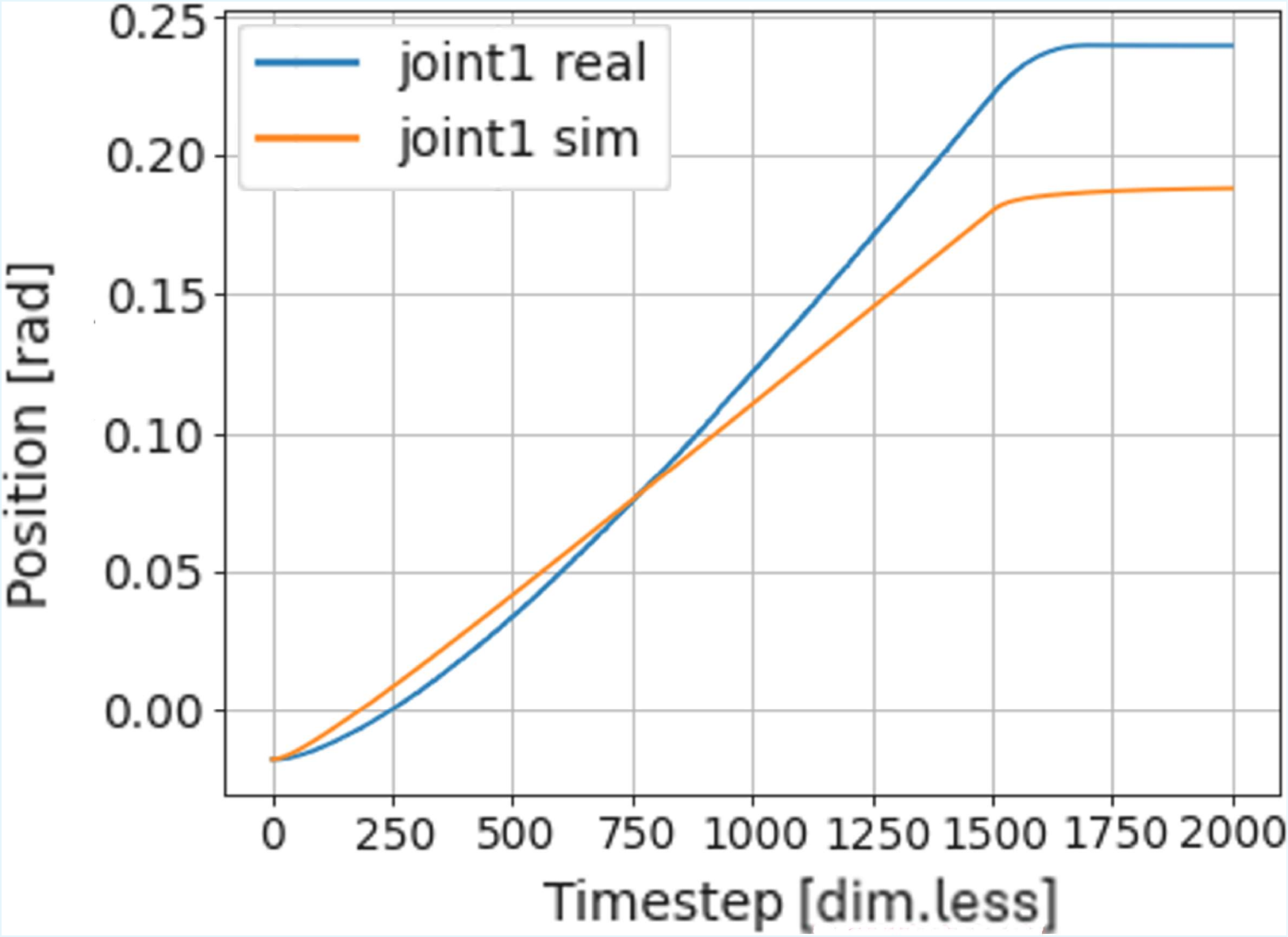} \label{fig:J1Tune}}
    \subfloat[]{\includegraphics[width=0.49\columnwidth]{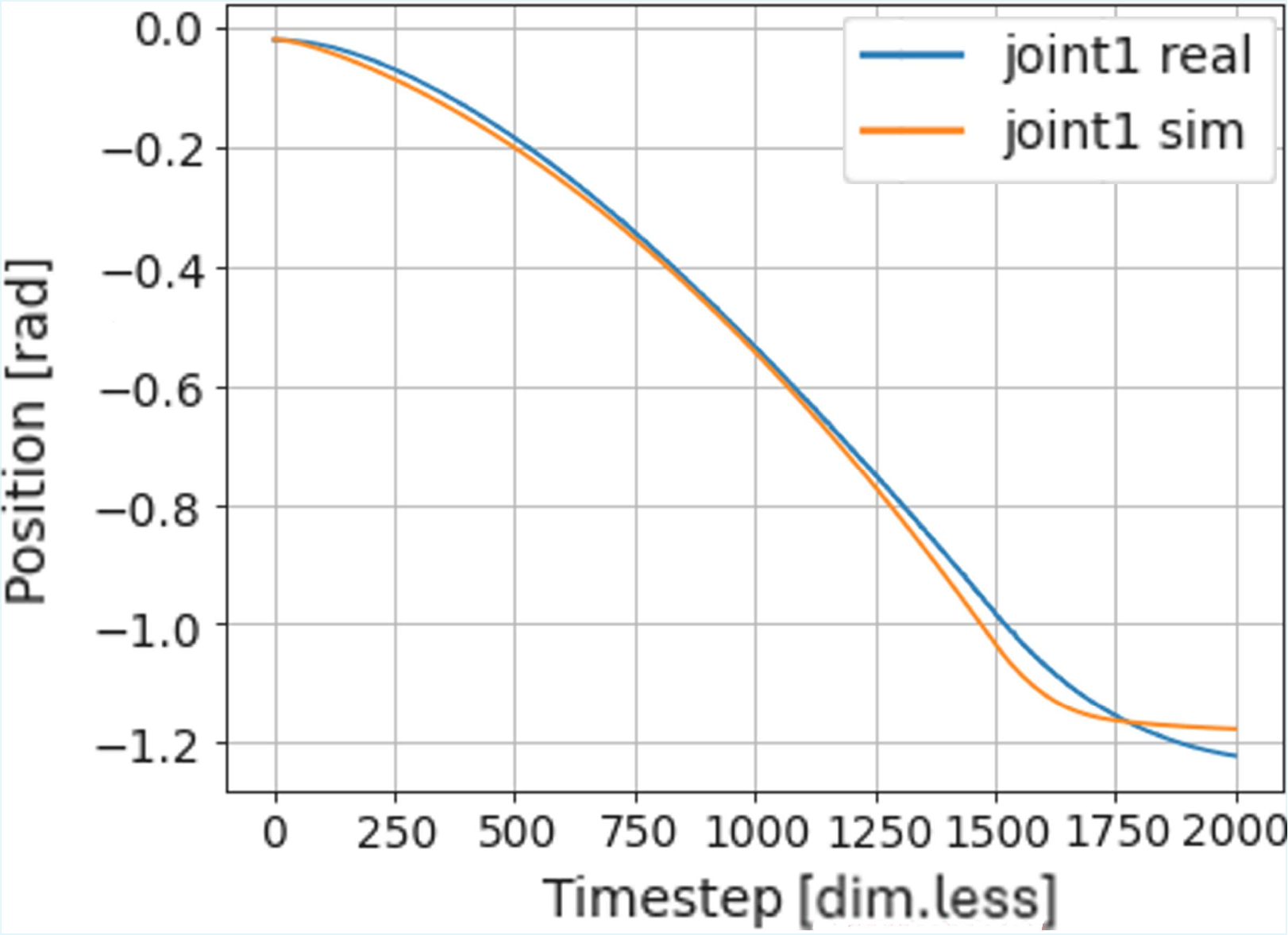} \label{fig:J2Tune}}
    \caption{\small Comparison of the Sim2Real RL model transfer for the Panda robot. 
    (a) Joint 1 torque comparison without parameter tuning. 
    (b) Joint 2 torque comparison without parameter tuning. 
    (c) Joint 1 torque comparison with parameter tuning. 
    (d) Joint 2 torque comparison with parameter tuning.}
    \label{fig:four_images}
    \vspace*{-4mm}
\end{figure}

Figs.\ref{fig:J1NoTune} and \ref{fig:J2NoTune} show some tests at the joint level without parameter calibration, inadequate gravity compensation, and unmodelled joint dynamics in load-bearing configurations.
To solve the gravity compensation issues, a torque offset was introduced for each joint, which aimed to make positive and negative trajectories visually similar. The offset values applied to each joint were determined through a manual adjustment process. The obtained torque values are described in Table \ref{tab:offset_values}.
% \begin{table}[H]
%   \centering
%   \begin{tabular}{c|c}
%     \hline
%     \textbf{Joint} & \textbf{Offset Value} \\
%     \hline
%     Joint 1 & -0.127 \\
%     Joint 2 & -0.725 \\
%     Joint 3 & 0.027 \\
%     Joint 4 & 0.53 \\
%     Joint 5 & 0 \\
%     Joint 6 & -0.06 \\
%     Joint 7 & 0 \\
%     \hline
%   \end{tabular}
%   \caption{Offset Values for Gravity Compensation and Kinetic Friction.}
%   \label{tab:offset_values}
% \end{table}

 Regarding the static friction, for accurate modelling, we applied a linearly increasing torque to each joint until movement was detected, establishing the minimum torque required to overcome static friction for both positive and negative directions. The resulting torque values, shown in Table \ref{tab:torque_values}, were used in simulation to replicate Real-World conditions by setting an action threshold; if the joint velocity remained within a small range and the applied torque was below the identified threshold, the action was nullified, simulating the static friction's hold on the joint.

After applying the Genetic Algorithm, the resulting inertia and friction parameters obtained are shown in Table \ref{tab:inertia_params} and \ref{tab:friction_params}.

 \begin{table}[!t]
 \vspace{7pt}
   \renewcommand{\arraystretch}{1.15}

  \centering
  \begin{tabular}{|c|c|c|c|c|c|c|c|}
    \hline
    \textbf{}       & \textbf{J1} & \textbf{J2} & \textbf{J3} & \textbf{J4} & \textbf{J5} & \textbf{J6} & \textbf{J7} \\
    \hline
    \textbf{Torque (Nm)} & -0.13 & -0.73 & 0.03 & 0.53 & 0 & -0.06 & 0 \\
    \hline
  \end{tabular}
  \caption{\small Offset Values for Gravity Compensation and static friction.}
  \label{tab:offset_values}
\end{table}

\begin{table}[!t]
\centering
  \renewcommand{\arraystretch}{1.15}

\begin{tabular}{|c@{\hspace{0.1cm}}|c@{\hspace{0.1cm}}|c@{\hspace{0.1cm}}|c@{\hspace{0.1cm}}|c@{\hspace{0.1cm}}|c@{\hspace{0.1cm}}|c@{\hspace{0.1cm}}|c|}
\hline
\textbf{Torque} & \textbf{J1} & \textbf{J2} & \textbf{J3} & \textbf{J4} & \textbf{J5} & \textbf{J6} & \textbf{J7} \\
\hline
\textbf{Pos. $\boldsymbol{\tau}$ (Nm)} & 0.65 & 0.88 & 0.51 & 0.61 & 0.44 & 0.40 & 0.32 \\
\textbf{Neg. $\boldsymbol{\tau}$ (Nm)} & -0.62 & -0.81 & -0.42 & -0.51 & -0.67 & -0.23 & -0.35 \\
\hline
\end{tabular}
\caption{\small Positive and Negative Friction Values for Each Joint.}
\label{tab:torque_values}
\end{table}

\begin{table*}[!t]
 \vspace{8pt}
   \renewcommand{\arraystretch}{1.15}

  \centering
  \begin{tabular}{|c|c|c|c|c|c|c|c|c|}
    \hline
    \textbf{} & \textbf{Inertia} & \textbf{Link 1} & \textbf{Link 2} & \textbf{Link 3} & \textbf{Link 4} & \textbf{Link 5} & \textbf{Link 6} & \textbf{Link 7} \\ \hline
    \multirow{6}{*}{\textbf{OLD}} & $I_{xx}$ & \num{0.70337} & \num{0.007962} & \num{0.03724} & \num{0.02585} & \num{0.03555} & \num{0.001964} & \num{0.01252} \\ %\cline{2-9}
                                 & $I_{yy}$ & \num{0.70661} & \num{0.02811}  & \num{0.03616} & \num{0.01955} & \num{0.02947} & \num{0.004354} & \num{0.01003} \\ %\cline{2-9}
                                 & $I_{zz}$ & \num{0.009117} & \num{0.025995} & \num{0.01083} & \num{0.02832} & \num{0.008627} & \num{0.005433} & \num{0.004815} \\ %\cline{2-9}
                                 & $I_{xy}$ & \num{-0.000139} & \num{-0.003925} & \num{-0.004761} & \num{0.007796} & \num{-0.002117} & \num{0.000109} & \num{-0.000428} \\ %\cline{2-9}
                                 & $I_{xz}$ & \num{0.006772} & \num{0.01025} & \num{-0.01140} & \num{-0.001332} & \num{-0.004037} & \num{-0.001158} & \num{-0.001196} \\ %\cline{2-9}
                                 & $I_{yz}$ & \num{0.019169} & \num{0.000704} & \num{-0.01281} & \num{0.008641} & \num{0.000229} & \num{0.000341} & \num{-0.000741} \\ \hline
    \multirow{6}{*}{\textbf{NEW}} & $I_{xx}$ & \num{0.5123} & \num{0.01089} & \num{0.04234} & \num{0.02525} & \num{0.04287} & \num{0.00148} & \num{0.01482} \\ %\cline{2-9}
                                 & $I_{yy}$ & \num{0.5147} & \num{0.03844} & \num{0.04110} & \num{0.01910} & \num{0.03554} & \num{0.00329} & \num{0.01188} \\ %\cline{2-9}
                                 & $I_{zz}$ & \num{0.0066}  & \num{0.03555} & \num{0.01231} & \num{0.02766} & \num{0.01040} & \num{0.00410} & \num{0.00570} \\ %\cline{2-9}
                                 & $I_{xy}$ & \num{-0.0001} & \num{-0.00537} & \num{-0.00541} & \num{0.007614} & \num{-0.00255} & \num{0.000082} & \num{-0.00051} \\ %\cline{2-9}
                                 & $I_{xz}$ & \num{0.0049}  & \num{0.01402} & \num{-0.01296} & \num{-0.001301} & \num{-0.00487} & \num{-0.00087} & \num{-0.00142} \\ %\cline{2-9}
                                 & $I_{yz}$ & \num{0.0140}  & \num{0.00096} & \num{-0.01456} & \num{0.00844}  & \num{0.00028} & \num{0.00026} & \num{-0.00088} \\ \hline
  \end{tabular}
  \caption{\small Inertia tensor components before (Old) and after (New) optimization for the 7 links.}
  \vspace{-3mm}
  \label{tab:inertia_params}
\end{table*}

\begin{table}[!t]
  \centering
  \setlength{\tabcolsep}{3.5pt}
  \renewcommand{\arraystretch}{1.15}
  \begin{tabular}{|c|>{\centering\arraybackslash}p{1.7cm}|c|c|c|c|c|c|c|}
    \hline
    \textbf{} & \textbf{Friction} & \textbf{J1} & \textbf{J2} & \textbf{J3} & \textbf{J4} & \textbf{J5} & \textbf{J6} & \textbf{J7} \\ \hline
    \multirow{3}{*}{\textbf{OLD}} & $\varphi_{1,j}$ & 0.55 & 0.87 & 0.64 & 1.28 & 0.84 & 0.30 & 0.56 \\ %\cline{2-9}
                                 & $\varphi_{2,j}$ & 5.12 & 9.07 & 10.14 & 5.59 & 8.35 & 17.13 & 10.34 \\ %\cline{2-9}
                                 & $\varphi_{3,j}$ & 0.04 & 0.03 & -0.05 & 0.04 & 0.03 & -0.02 & 0.01 \\ \hline
    \multirow{3}{*}{\textbf{NEW}} & $\varphi_{1,j}$ & 3.40 & 1.98 & 2.51 & 2.44 & 2.60 & 1.19 & 1.55 \\ %\cline{2-9}
                                 & $\varphi_{2,j}$ & 9.67 & 28.67 & 9.83 & 18.78 & 4.19 & 5.75 & 1.70 \\ %\cline{2-9}
                                 & $\varphi_{3,j}$ & 0.01 & 0.01 & -0.01 & 0.01 & 0.01 & -0.01 & 0.01 \\ \hline
  \end{tabular}
  \caption{\small Friction parameters before (Old) and after (New) optimization.}
  \label{tab:friction_params}
\end{table}

\subsection{Sim2Sim}\label{ss:Sim2Sim}
\begin{figure}[t]
    \centering
    \subfloat[]{
    \includegraphics[width=0.48\columnwidth]{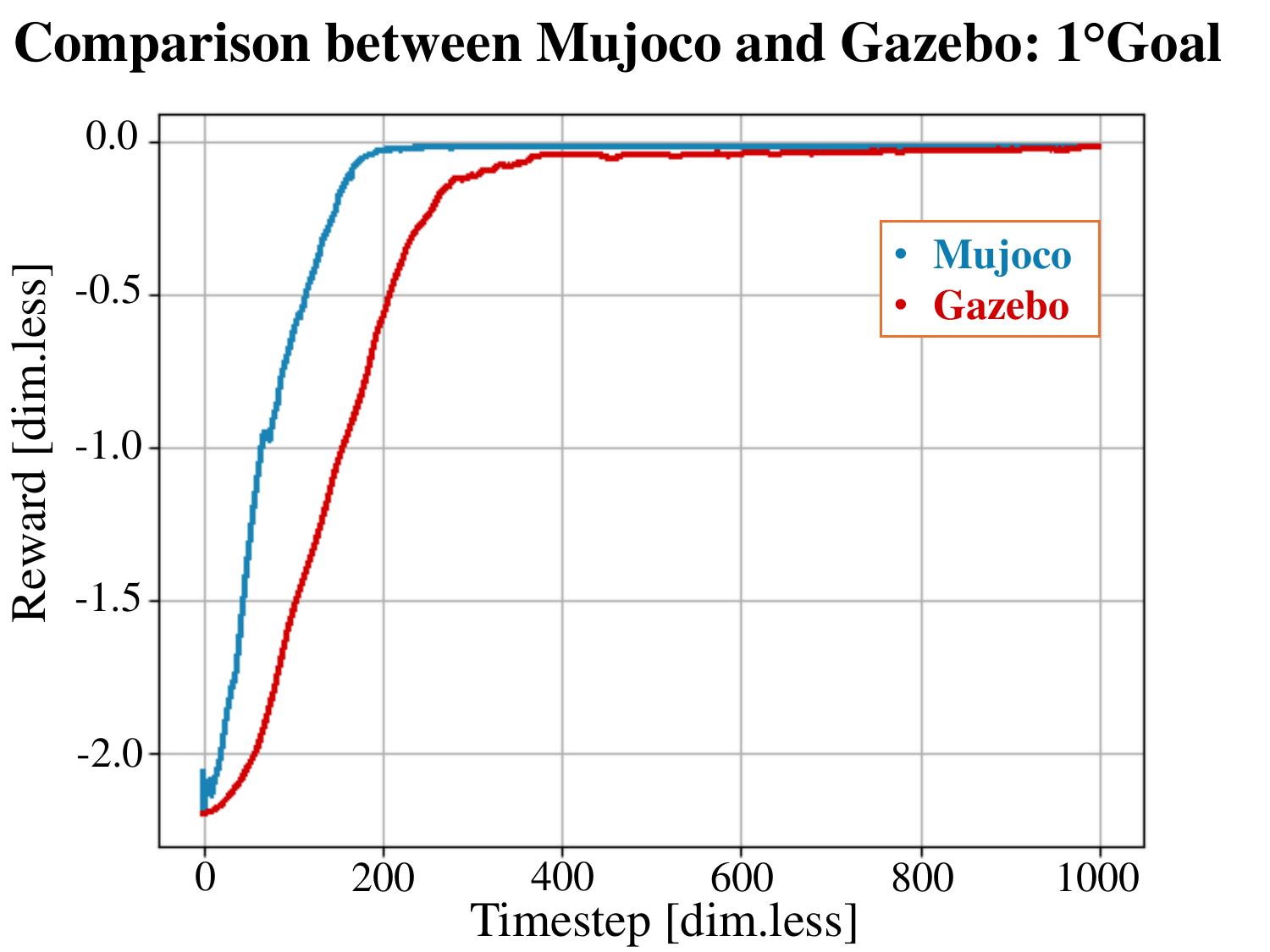}
    }\label{subfig:MG1g}\subfloat[]{
    \includegraphics[width=0.48\columnwidth]{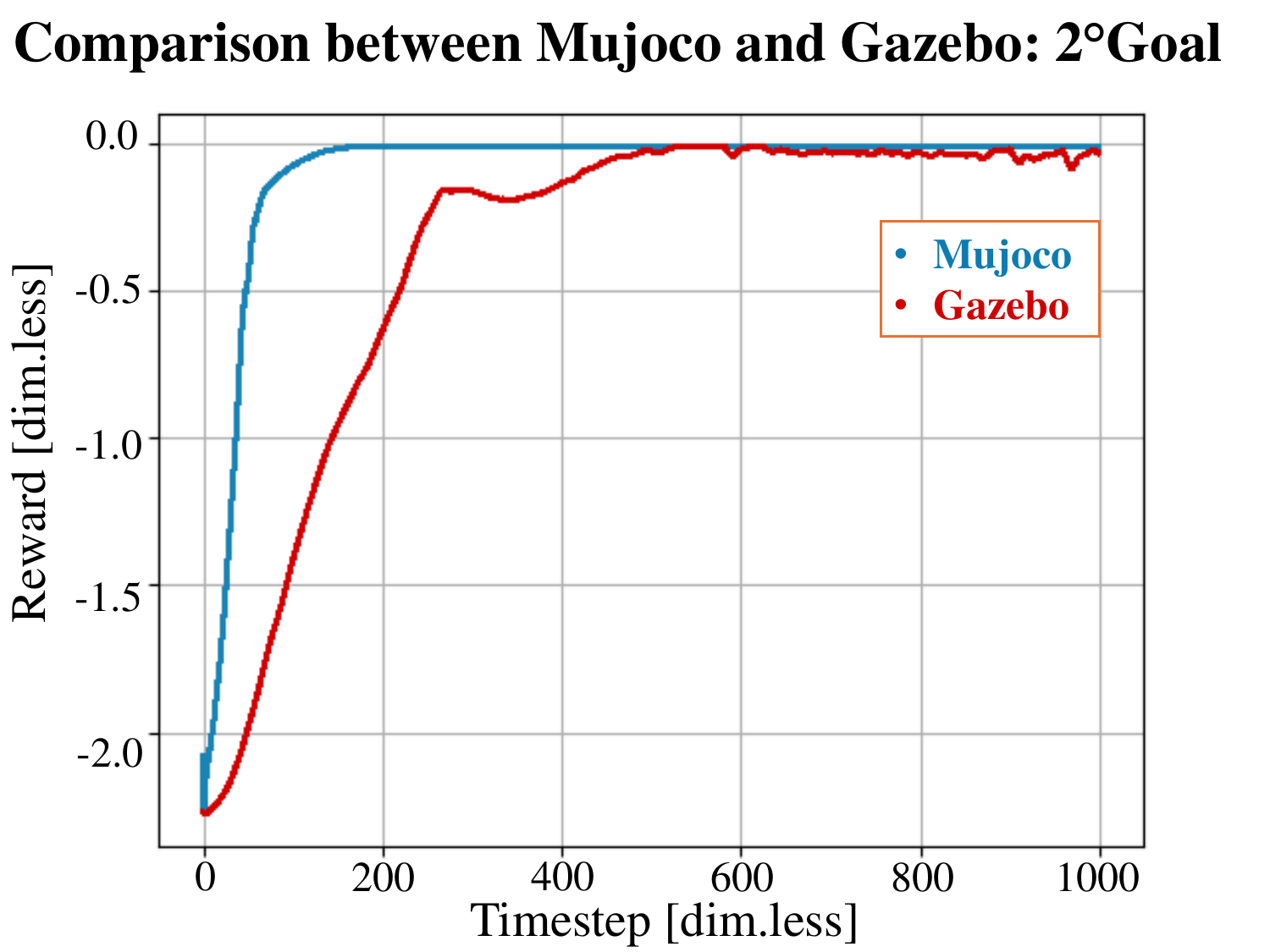}
    }\label{subfig:MG2g}

    \subfloat[]{
    \includegraphics[width=0.48\columnwidth]{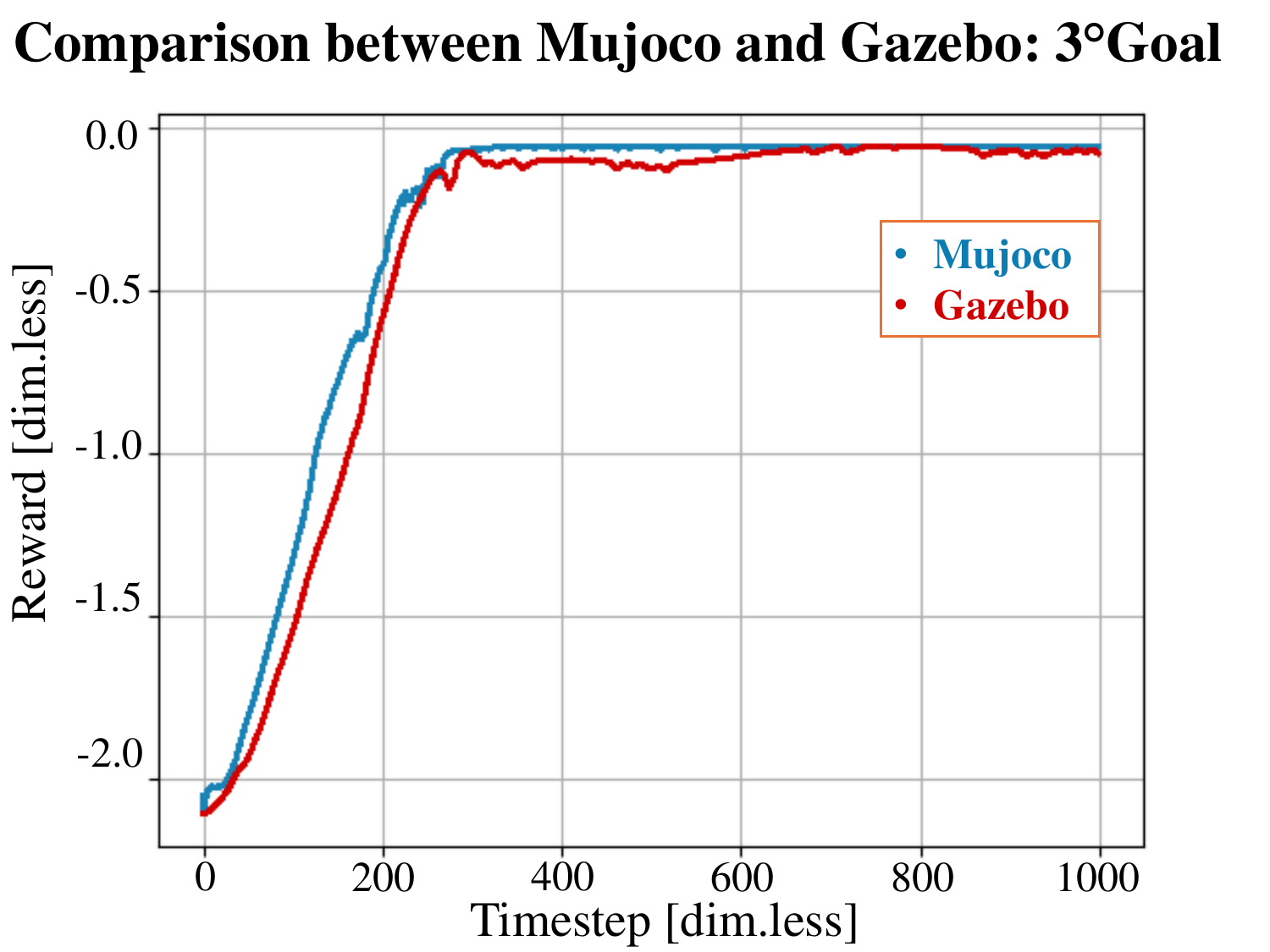}
    }\label{subfig:MG3g}\subfloat[]{
    \includegraphics[width=0.48\columnwidth]{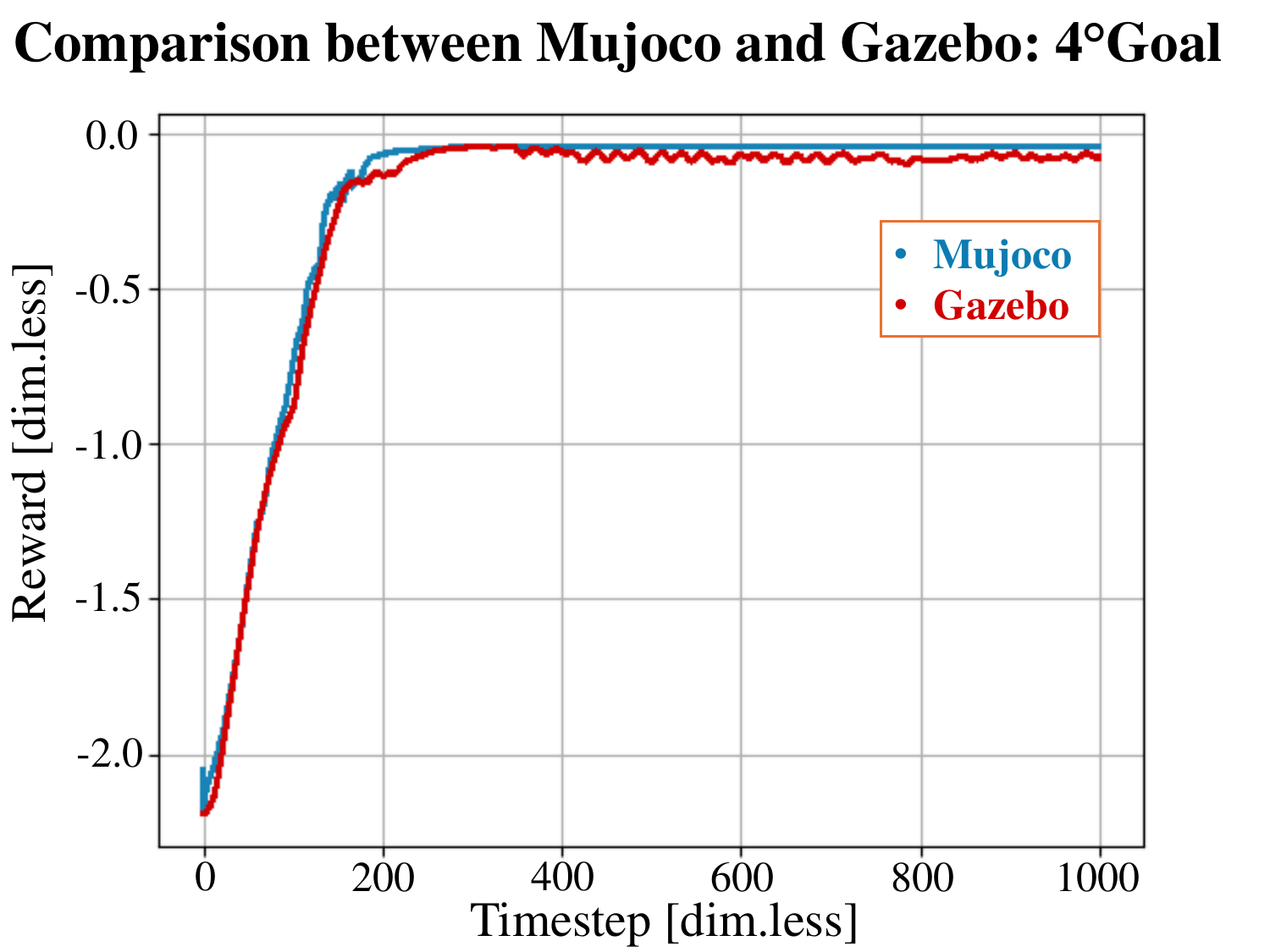}
    }\label{subfig:MG4g}
    \caption{\small Comparison of the TQC agent's performance in the MuJoCo and Gazebo simulators. The plots display the rewards obtained across four random goal scenarios, emphasizing the performance variations resulting from the distinct dynamics and properties of each simulation environment.}
    \label{fig:graphsMG}
\end{figure}

\begin{figure}[t]
    \centering
    \includegraphics[width = 0.95\columnwidth]{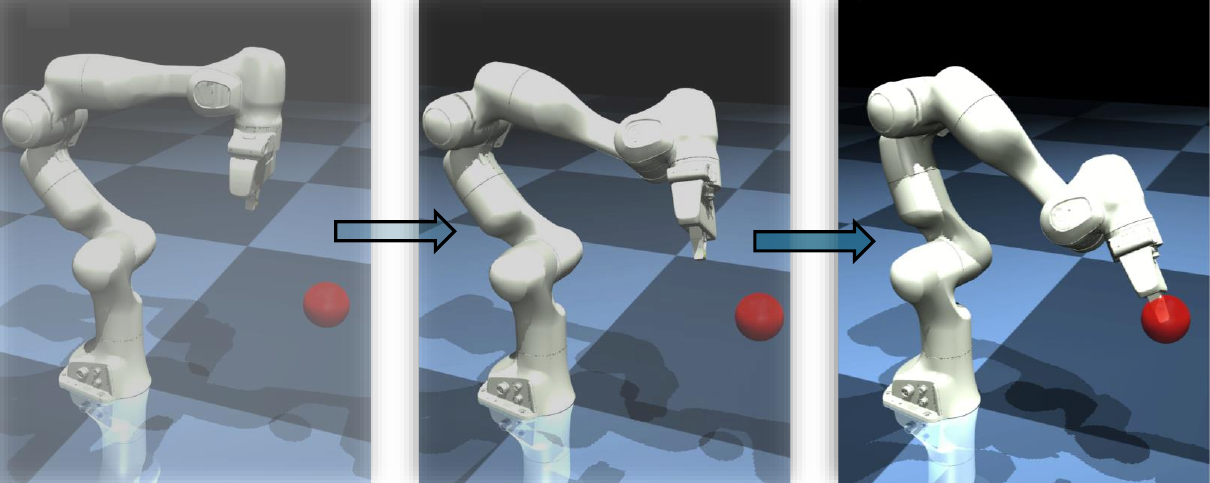}
    \caption{\small Visualization of the Panda robot reaching the goal in the MuJoCo simulator. The image depicts the simulated trajectory and final pose, showcasing task completion in the virtual environment.}
    \label{fig:PandaMujoco}
\end{figure}

\begin{figure}[t]
    \centering
    \includegraphics[width = 0.95\columnwidth]{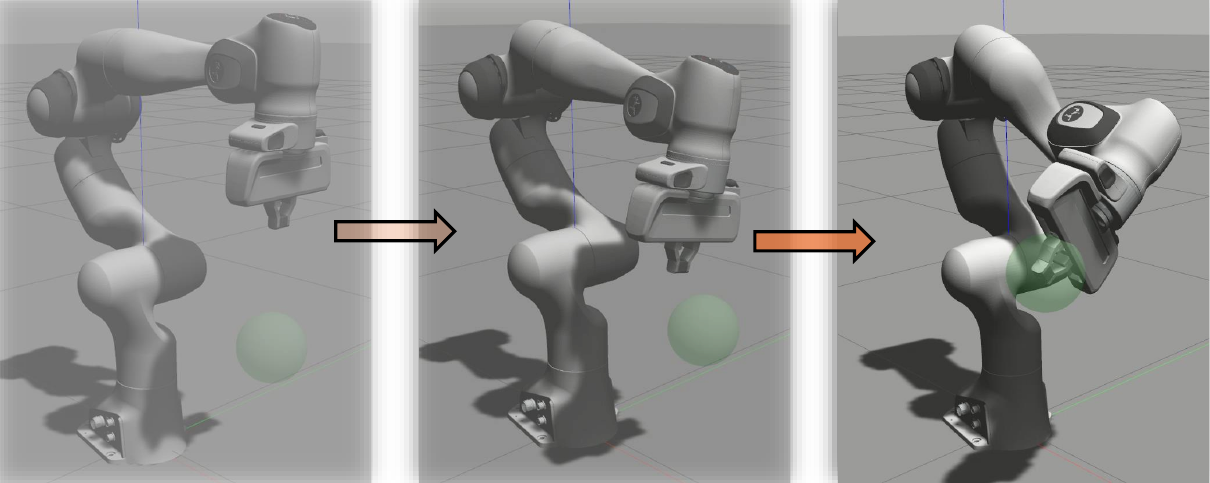}
    \caption{\small Visualization of the Panda robot reaching the goal position in the Gazebo simulator. The image highlights the robot's path and final pose, demonstrating successful performance in the simulated setting.}
    \label{fig:PandaGazebo}
    \vspace{-5mm}
\end{figure}
In the simulation setup, the TQC agent is trained in MuJoCo to reach random target points across diverse workspace coordinates, since MuJoCo provides higher simulation speed and is therefore better suited for rapid policy training.
Before deploying the agent in the Real-World, we used the Gazebo simulator with different Panda model parameters inside the training domain, in order to test the behaviour of the robot with different targets.

The primary task for the agent is to guide the robotic arm to reach random target points within its workspace, simulating Real-World challenges where the arm must adapt to varying target positions. The agent’s adaptability is thoroughly tested by assigning random target coordinates, requiring it to handle diverse end-effector locations. In addition to target variability, DR was applied to the simulation by perturbing all 21 joint-specific friction parameters~$\varphi$ within a range of $\pm10\%$, and all 42 elements of the joint inertia matrices within a range of $\pm5\%$ of their previously estimated values (see Table \ref{tab:inertia_params}). These perturbation ranges were selected based on preliminary empirical tests to balance variability with stability. Larger perturbations led to unstable training, while smaller ones reduced the agent's generalization capability. This setup not only validates the algorithm’s robustness but also highlights its practical applicability in dynamic and uncertain scenarios.

\subsection{Sim2Real}\label{ss:Sim2Real}

\begin{figure}[t]
    \centering    \vspace{2mm}    \includegraphics[width = \columnwidth]{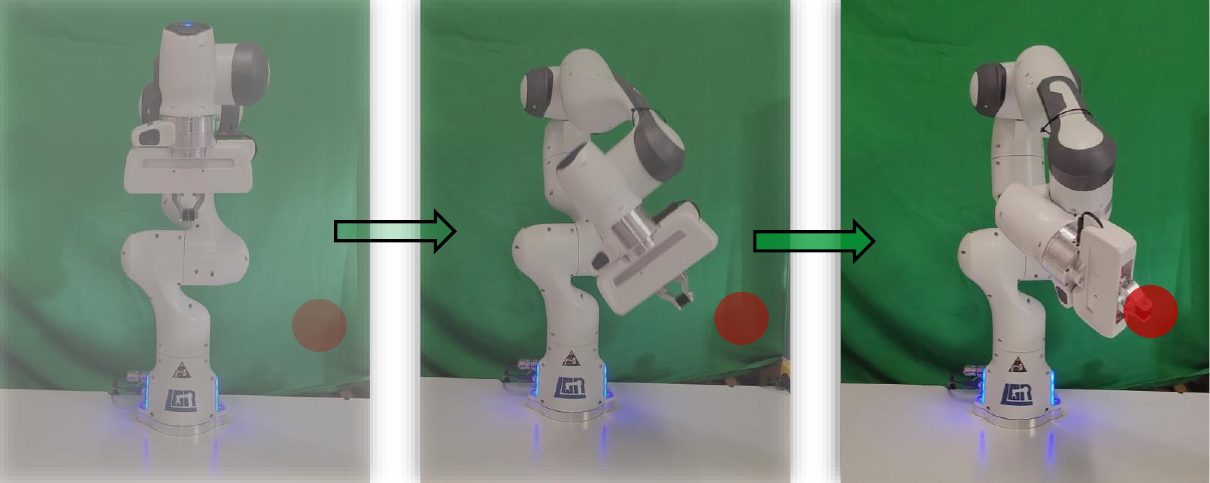}
    \caption{\small Visualization of the Panda robot reaching the goal position in the Real-World environment. The image captures the robot's trajectory and final pose, emphasizing successful task execution in a physical setup.}
    \label{fig:PandaReal}
\end{figure}

\begin{figure}[t]
    \centering
    \subfloat[]{
    \includegraphics[width=0.48\columnwidth]{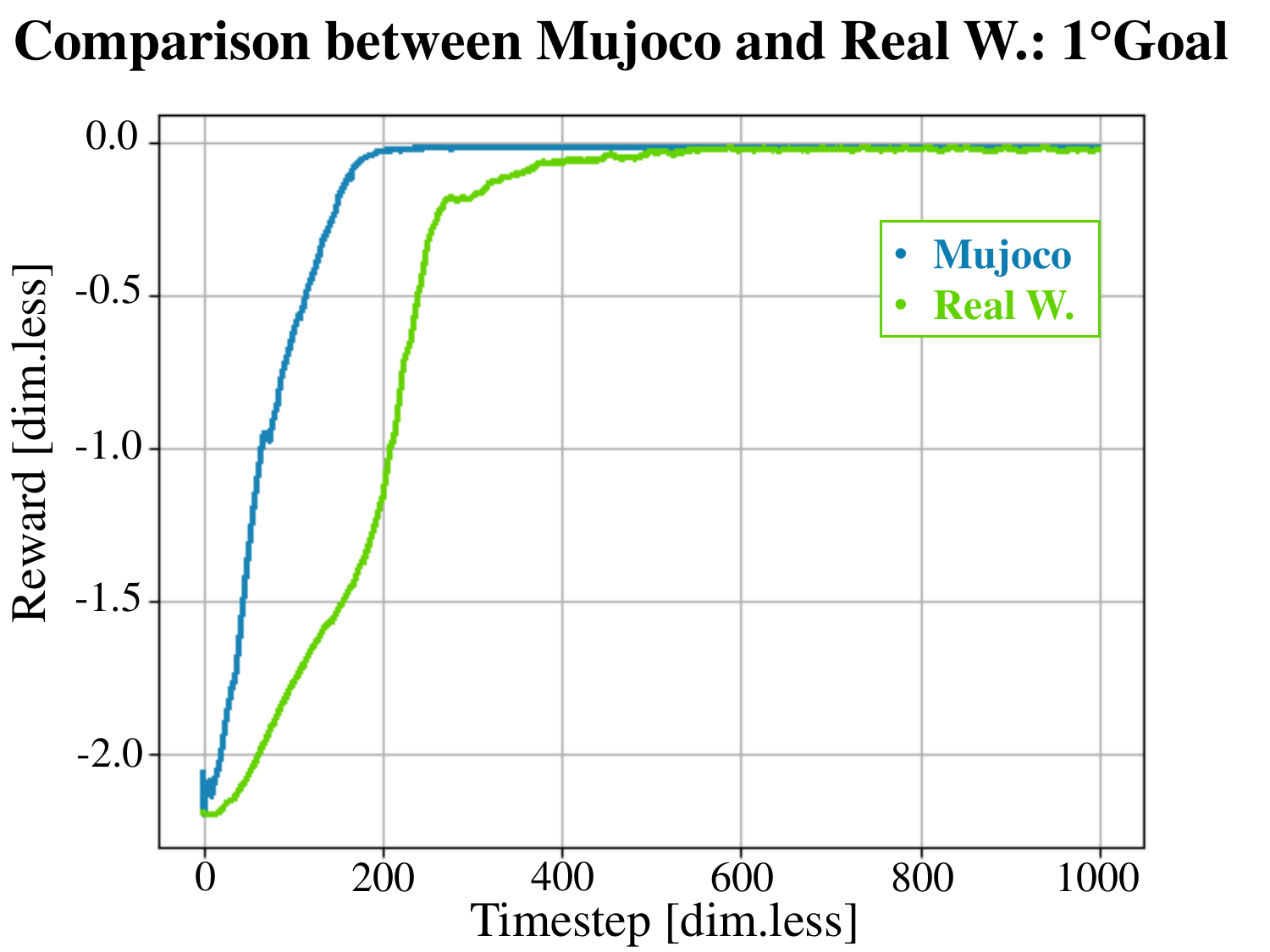}
    }\label{subfig:MR1g}\subfloat[]{
    \includegraphics[width=0.48\columnwidth]{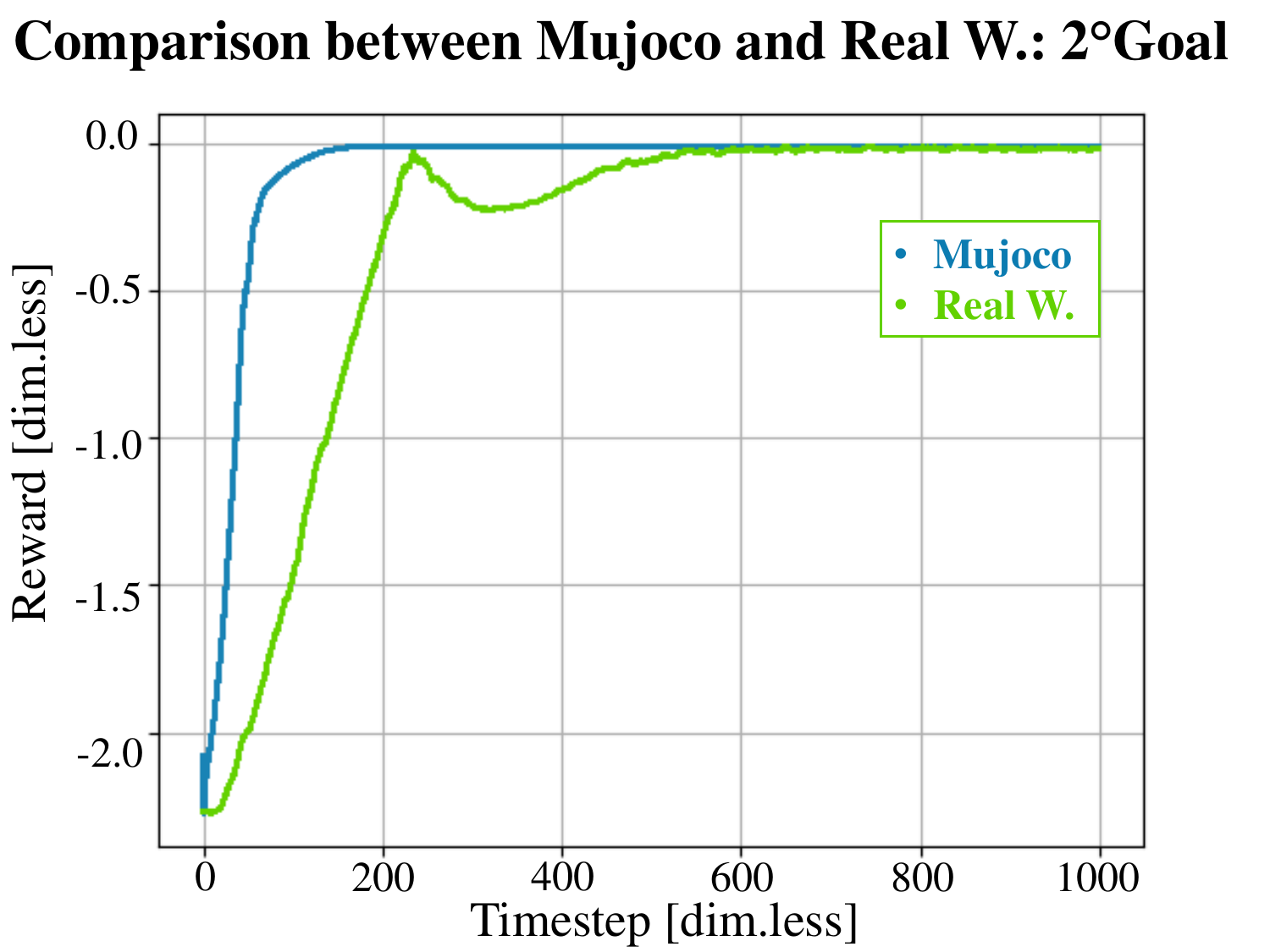}
    }\label{subfig:MR2g}

    \subfloat[]{
    \includegraphics[width=0.48\columnwidth]{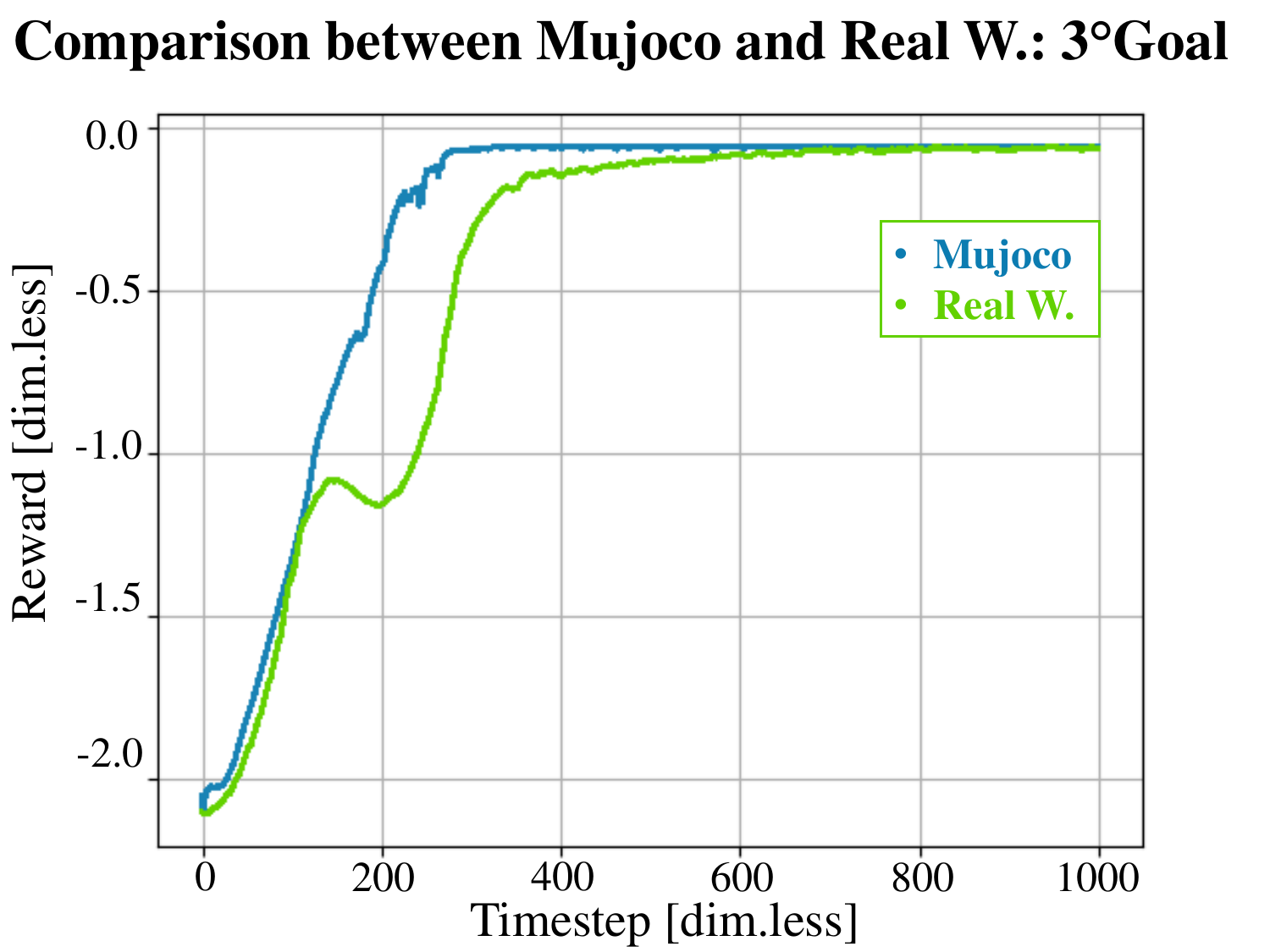}
    }\label{subfig:MR3g}\subfloat[]{
    \includegraphics[width=0.48\columnwidth]{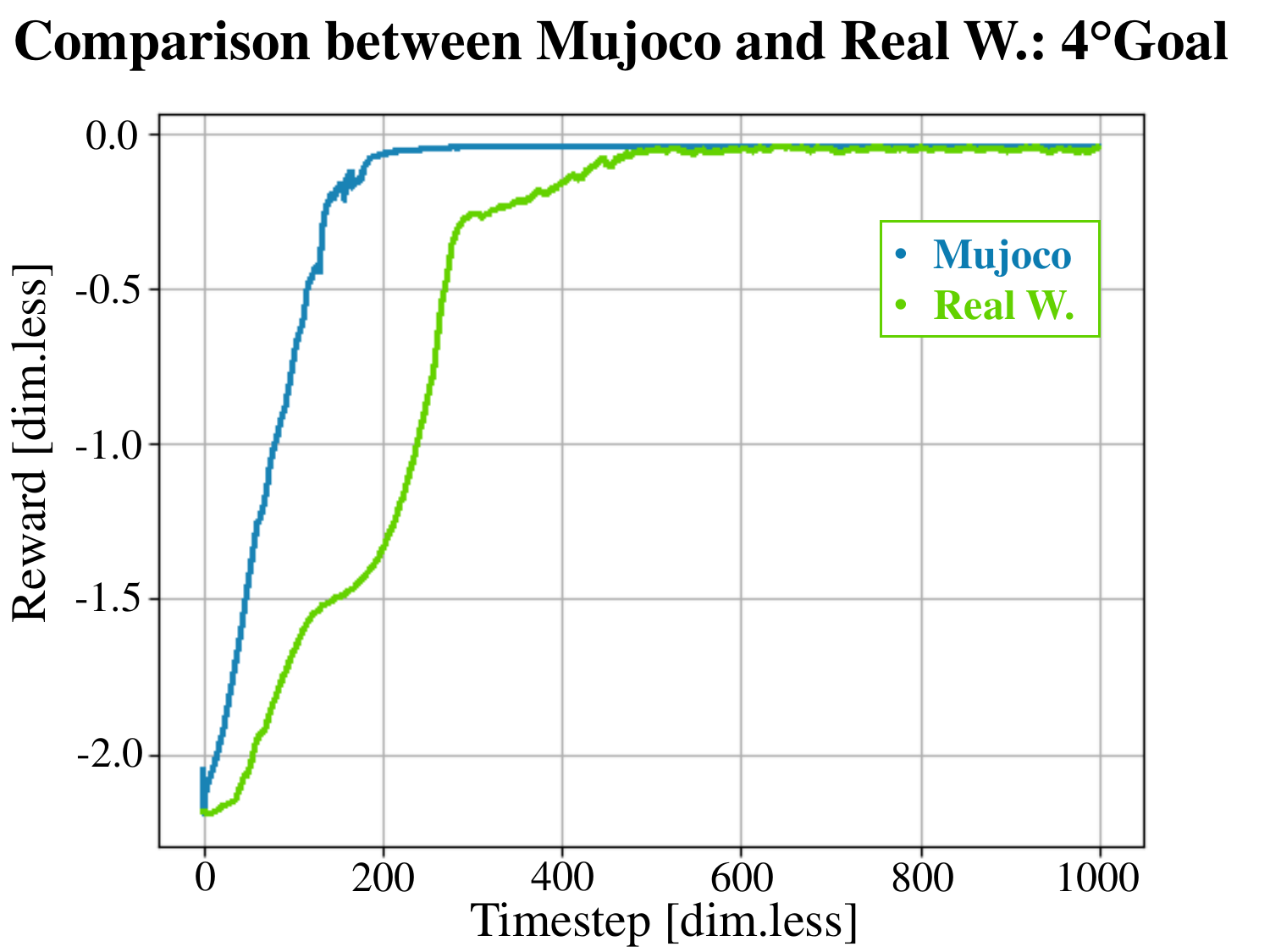}
    }\label{subfig:MR4g}
    \caption{\small Comparison between MuJoCo and Real-World for four random goals. MuJoCo shows smoother transitions while Gazebo introduces disturbances mimicking physical inconsistencies, serving as an intermediate test-bed before real deployment.}
    \label{fig:graphsMR}
    \vspace{-5mm}
\end{figure}

In transferring our RL agent from simulation to the Real-World environment, we focused on ensuring robust performance by preparing the agent for varied conditions during training. One technique we used to support this transition was DR, where simulated physical parameters such as friction and inertia were randomized. This method helps bridge the "reality gap" by training the agent across a broad spectrum of conditions, making it more adaptable to Real-World uncertainties.

After training, the agent was deployed on the physical Franka Emika Panda robot, achieving high reward scores and demonstrating reliable control. The agent's exposure to domain-randomized parameters during simulation enabled it to generalize effectively to Real-World conditions, requiring minimal adjustments. As shown in the comparison between MuJoCo simulation and Real-World trials, the plots illustrating the Real-World evaluation indicate that the model trained with DR performs remarkably well in the actual physical environment. The end effector consistently approaches and maintains its position near the randomized goal. Notably, some minor uncertainties are observed in reaching the target, as depicted in Fig.~\ref{fig:graphsMR} (b) and Fig.~\ref{fig:graphsMR} (c). Despite these slight deviations, the overall performance demonstrates the effectiveness of the trained model in a Real-World setting.

The evaluation across different environments shows the generalizability and robustness of the trained model, allowing possible applications in Real-World settings.

\section{Conclusions} \label{s:Conclusions}
In this work, we presented a \textit{Real2Sim2Real} pipeline for RL in torque-controlled robots, aimed at improving Sim2Real transfer under low-level actuation. By combining trajectory matching, DR, and genetic optimization of simulation parameters, we reduced the reality gap on a 7-DOF Franka Emika Panda.

Our results show that physics-aware system identification, especially relying on tuning a custom function of friction and inertia instead of relying on simulation friction parameters, increases simulation fidelity, enabling policies trained entirely in simulation to transfer to the real robot. Among the evaluated RL methods, TQC provided the most reliable performance. The proposed reward, balancing task accuracy and smooth actuation, led to stable and physically plausible motions.

We validated the framework not only in Sim2Real, but also in \textit{Sim2Sim} across two simulators (MuJoCo and Gazebo), highlighting that the proposed calibration and training procedure is effective beyond a single simulation domain. Real-robot experiments further confirmed robustness under practical disturbances (e.g., backlash, cable drag, and unmodelled delays).

Future directions include extending to contact-rich manipulation tasks, investigating online system identification, generalizing to other robot platforms, and integrating safety-aware RL and compliance control.

\bibliographystyle{IEEEtran}
\bibliography{biblio}

\end{document}